\documentclass[lettersize,journal]{IEEEtran}
\usepackage{amsmath,amsfonts}
\usepackage{algorithmic}
\usepackage{algorithm}
\usepackage{array}
\usepackage[caption=false,font=normalsize,labelfont=sf,textfont=sf]{subfig}
\usepackage{textcomp}
\usepackage{stfloats}
\usepackage{url}
\usepackage{verbatim}
\usepackage{graphicx}
\usepackage{cite}
\usepackage{multirow}
\usepackage{color}
\usepackage{changepage}
\usepackage{float}

\usepackage{amssymb}
\usepackage{multirow}
\usepackage{bbding}
\usepackage{booktabs}
\usepackage{caption}
\usepackage{bm}
\usepackage{bbm}
\usepackage{tabu}
\usepackage{stfloats}
\usepackage{makecell}
\usepackage{ltxtable}
\usepackage{setspace}
\usepackage[figuresright]{rotating}
\usepackage{longtable}
\usepackage{threeparttable}
\usepackage{tabularx}
\usepackage{hyperref}
\usepackage{ragged2e}
\usepackage{xcolor}
\begin{document}

\title{Few-Shot Class-Incremental Learning with Prior Knowledge}


\author{Wenhao Jiang, {\it Student Member, IEEE}, Duo Li$^*$, Menghan Hu$^*$, {\it Senior Member, IEEE, Guangtao Zhai, Senior Member, IEEE, Xiaokang Yang, Fellow, IEEE, Xiao-Ping Zhang, Fellow, IEEE} 

\thanks{This work is sponsored by the National Natural Science Foundation of China (No. 62371189).\\
\indent Wenhao Jiang, and Menghan Hu are with the Shanghai Key Laboratory of
Multidimensional Information Processing, School of Communication and
Electronic Engineering, East China Normal University, Shanghai 200241,
China. \\
\indent Duo Li is with the Kargobot of DiDi, Shanghai 201210, China. \\
\indent Guangtao Zhai, and Xiaokang Yang are with the Institute of Image
Communication and Network Engineering, Shanghai Jiao Tong University
Shanghai, 200240, China. \\
\indent Xiao-Ping Zhang is with Tsinghua Berkeley Shenzhen Institute, Shenzhen,
China and the Department of Electrical, Computer and Biomedical Engineering, Toronto Metropolitan University, ON M5B 2K3, Canada.\\
\indent $^*$Corresponding authors: Duo Li; Menghan Hu} }
        



\maketitle

\begin{abstract}
To tackle the issues of catastrophic forgetting and overfitting in few-shot class-incremental learning (FSCIL), previous work has primarily concentrated on preserving the memory of old knowledge during the incremental phase. The role of pre-trained model in shaping the effectiveness of incremental learning is frequently underestimated in these studies. Therefore, to enhance the generalization ability of the pre-trained model, we propose Learning with Prior Knowledge (LwPK) by introducing nearly free prior knowledge from a few unlabeled data of subsequent incremental classes. We cluster unlabeled incremental class samples to produce pseudo-labels, then jointly train these with labeled base class samples, effectively allocating embedding space for both old and new class data. Experimental results indicate that LwPK effectively enhances the model resilience against catastrophic forgetting, with theoretical analysis based on empirical risk minimization and class distance measurement corroborating its operational principles. The source code of LwPK is publicly available at: \url{https://github.com/StevenJ308/LwPK}. 
\end{abstract}
\begin{IEEEkeywords}
few-shot learning, class-incremental learning, prior knowledge
\end{IEEEkeywords}

\section{Introduction}
\IEEEPARstart{I}n recent years, deep learning has made breakthroughs in various vision tasks. In real-world scenarios, data often comes in sequentially as streams, which poses challenges for us to train neural networks: (I) It is expensive to retrain the network when new data is loaded in, and in specific cases, the old data cannot be reused; (II) training only on new data can make the neural network's ability to process old knowledge drop dramatically, called catastrophic
    \begin{figure}[htb]
    \centering
    \includegraphics[height=6.2cm,width=8.5cm]{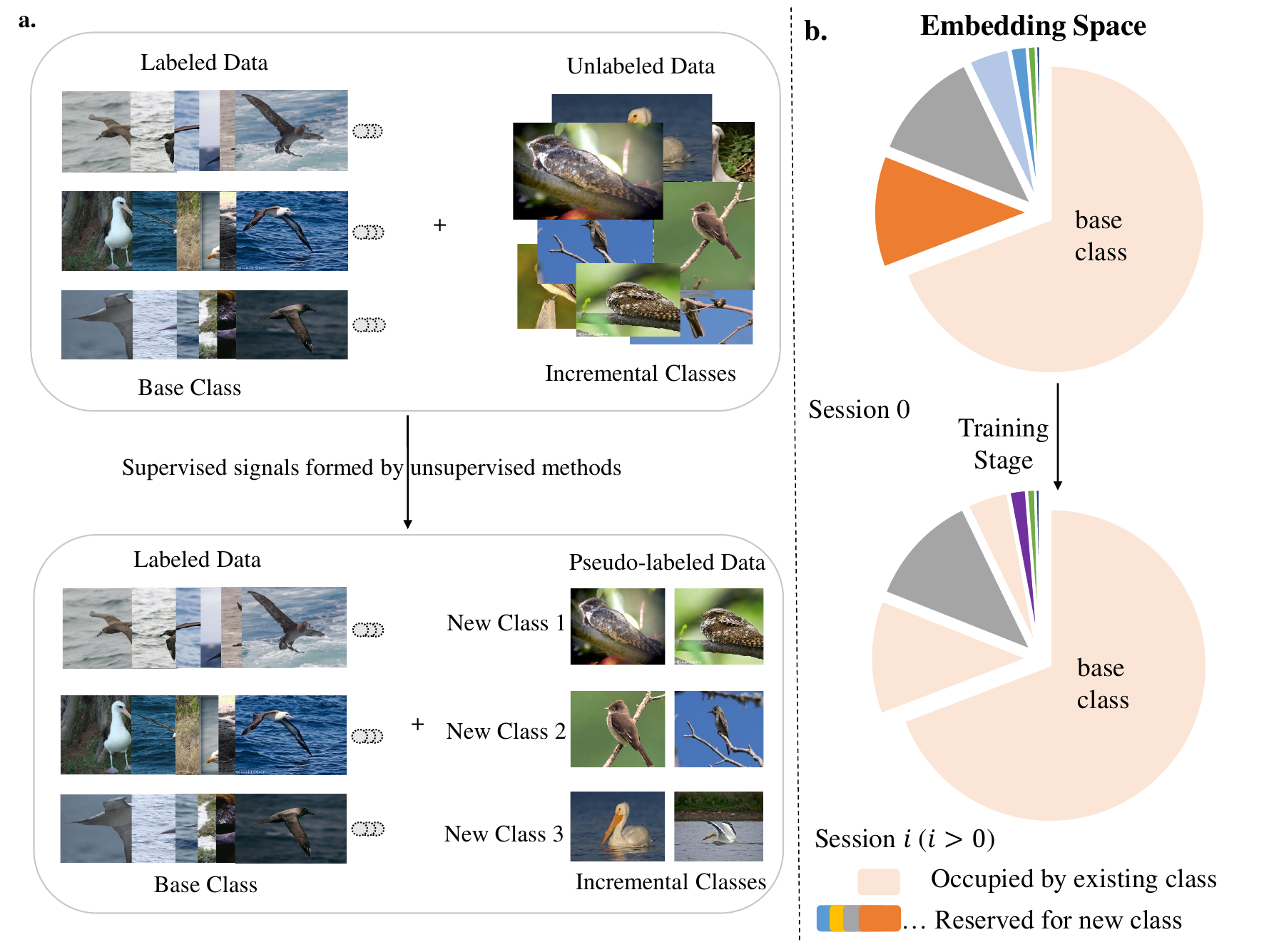}
    \caption{Illustration of the operations we take to build a model with prior knowledge: a) the training data consists of labeled base class data and unlabeled incremental class data, and we provide supervised signals to the unlabeled data by unsupervised methods. b) the embedding space occupied by existing class data and reserved for new class data, thereby demonstrating the influence of the incremental class data on the construction of a pre-trained model.}
    \label{fig:fig1}
    \end{figure}
forgetting. Therefore, incremental learning is needed to help solve these problems.

   
Current class incremental learning focuses more on the case when the number of new class samples is large enough. In real life, the label is often harder to obtain, especially for rare resources or data that need to focus on privacy protection. This led to the emergence of the few-shot class-incremental learning (FSCIL) \cite{tao2020topology}, \cite{ravi2016optimization}, \cite{finn2017model}, \cite{snell2017prototypical}, \cite{sung2018learning}, \cite{10018890}. The fundamental paradigm of FSCIL involves pre-training a model on a comprehensive labeled base class dataset. Subsequent data arrives in the form of a stream of \(N\)-way \(K\)-shot, necessitating the model to minimize the forgetting of previous knowledge while assimilating new information.

In the face of challenges associated with learning and convergence in few-shot data streams, conventional attention is often directed toward optimizing the efficient utilization of limited samples. However, there is a tendency to overlook the valuable semantic information that could potentially reside within unlabeled data. Leveraging these unlabeled data and integrating them into the initial model training process can furnish specific feature priors for subsequent incremental learning. Simultaneously, this approach facilitates the acquisition of shared patterns, thereby augmenting the model's generalization capabilities.

Recently, there has been research leveraging readily available unlabeled data to enhance model performance in the FSCIL task \cite{cui2021semi}, \cite{cui2022uncertainty},\cite{cui2023uncertainty}. They have achieved good results using a semi-supervised learning paradigm and utilizing knowledge distillation. This work can be categorized as semi-supervised few-shot incremental learning (Semi-FSCIL).
    
    
    Based on the current state of research, we intend to leverage unlabeled data to incorporate prior knowledge during the pre-training phase, setting the stage for subsequent incremental learning, as illustrated in Figure \ref{fig:fig1}. 
    Semi-supervised learning has demonstrated its effectiveness in enhancing models with unlabeled data when labeled data is limited \cite{berthelot2019mixmatch}, \cite{sohn2020fixmatch}. In our approach, we leverage unlabeled data in a manner that differs from traditional semi-supervised approaches. The unlabeled data is not used as labeled cooperation objects simultaneously with labeled data; instead, it is introduced at the very beginning as a form of data prior. Previous work \cite{zhou2022forward} explains the advantages of forward compatibility in this context, while Germain \textit{et al}. \cite{germain2009pac} demonstrates the validity of the priori from a theoretical perspective. They do not extract valid information from easily available unlabeled data to enhance modeling capabilities. We initiate with novel concepts, proposing the idea of incorporating knowledge priors in the pre-training.

    Wu \textit{et al.} \cite{wu2022class} proposes a two-stage algorithm that first builds a basic model with strong representational capabilities and adapts it to new data by fine-tuning. Their approach requires pre-training on a very large dataset and is not well-suited for few-shot tasks.
    In contrast, by adding weaker prior knowledge as an intervention in the pre-trained phase, we can improve the extraction and recognition of new knowledge in the incremental phase by learning as many features as possible that may subsequently appear as new classes; at the same time, fewer adjustments to the model parameters can better maintain the model's memory of old knowledge. LwPK needs not to train a large amount of base data in the pre-trained phase but only needs to add some unlabeled images and assign pseudo-labels using unsupervised clustering. By training these data jointly, the pre-trained model with prior knowledge can be constructed. To assign supervisory signals to unlabeled data, we utilize a deep clustering scheme that performs clustering and pseudo-label assignment on a semantic-based basis by learning feature representations that are more suitable for clustering \cite{yang2017towards}, \cite{tao2021clustering}.

    The contributions of this study mainly include:
\begin{itemize}
    \item Propose LwPK to mitigate catastrophic forgetting and overfitting in FSCIL using unlabeled samples with pseudo-labels generated by clustering algorithm for joint training with base class data.
    \item Experimentally confirm the effectiveness of LwPK for FSCIL through extensive experiments on CIFAR100 \cite{krizhevsky2009learning}, CUB200 \cite{wah2011caltech}, and {\it mini}ImageNet \cite{vinyals2016matching}.     
    \item Theoretically substantiate the efficacy of LwPK through empirical risk minimization and class distance analysis, aligning with experimental findings and showcasing its robust approach in FSCIL.
\end{itemize}
\section{Related Work}
    \textbf{Few-shot Learning} FSL refers to learning knowledge from few-shot training samples, whose data volume is often in single digits \cite{ravi2016optimization}, \cite{finn2017model}, \cite{snell2017prototypical}, \cite{sung2018learning}, \cite{vinyals2016matching}, \cite{jamal2019task}, \cite{hariharan2017low}, \cite{kwitt2016one}, \cite{9136838}. The primary challenge in FSL lies in the unreliability of empirical risk minimization due to very limited sample sizes. Presently, mainstream FSL solutions can be categorized into three types: data augmentation-based approaches \cite{hariharan2017low}, \cite{kwitt2016one}, optimization-based approaches \cite{ravi2016optimization}, \cite{finn2017model}, and metric-based approaches \cite{snell2017prototypical}, \cite{sung2018learning}. The data augmentation-based approach is straightforward, as it aims to expand the dataset to provide the model with sufficient knowledge to optimize its parameters. Optimization-based approaches aim at designing networks or algorithms to allow models to be adapted quickly on limited data. The metric-based approaches extract image features by a trained model and select an appropriate metric for image matching. Commonly used metrics include distance metrics, cosine similarity, Deepemd \cite{zhang2020deepemd}, etc. Our work combines elements of both data augmentation and measurement.
    
    \textbf{Class-incremental Learning} CIL is employed to mitigate the issue of catastrophic forgetting induced by the introduction of new classes in data stream \cite{li2017learning}, \cite{rebuffi2017icarl}, \cite{castro2018end}, \cite{hou2019learning}, \cite{chaudhry2019continual}, \cite{rolnick2019experience}, \cite{bang2021rainbow}, \cite{abati2020conditional}, \cite{yan2021dynamically}, \cite{10002397}. Current techniques fall into three main categories: regularization-based methods \cite{li2017learning}, \cite{hu2021distilling}, \cite{rebuffi2017icarl}, replay-based methods\cite{rebuffi2017icarl}, \cite{castro2018end}, \cite{van2020brain} and dynamic network-based methods \cite{abati2020conditional}, \cite{yoon2017lifelong}, \cite{yan2021dynamically}. Regularization methods aim to mitigate the forgetting of old knowledge by introducing constraints into the loss function of the new task, with distillation loss being a common constraint. The replay approach involves retaining some old data to create an example set alongside the new dataset. The example set can be optimized using strategies like herd selection and balancing. Dynamic networks allocate specific parameters to different tasks, allowing the avoidance of forgetting old knowledge when new data arrives by expanding the feature extraction model or classification head. Some recent work has also emerged that addresses CIL from the perspective of pre-trained models. It has been proven that strong pre-trained models can achieve better performance in downstream tasks \cite{yosinski2014transferable}. Some recent work has also emerged that addresses CIL from the perspective of pre-trained models. In \cite{wu2022class}, Wu \textit{et al.} trains 800 base classes in the pre-training phase to improve the representational ability of the model, while a large amount of base class knowledge provided a more general feature extractor for the subsequent incremental phase.
    
    \textbf{Few-shot Class-incremental Learning} FSCIL is recently proposed for solving the problem of scenarios with a small number of samples in CIL. Tao \textit{et al.} \cite{tao2020few} proposes a Topology-Preserving Knowledge InCrementer (TOPIC) framework to address the problem of forgetting old knowledge and learning new samples by using neural gases. In Deep-EMD \cite{zhang2020deepemd}, a new perspective from optimal matching between image regions is used to develop a shot-less image classification method. To adopt this approach, Zhang \textit{et al.} designs a cross-referencing mechanism that effectively mitigates the adverse effects of background clutter and large variations in intra-class appearance. In semantic-aware knowledge distillation for FSCIL \cite{cheraghian2021semantic}, Cheraghian \textit{et al.} introduces word vectors of new classes and uses embedding and attention modules to solve the semantic mapping problem for completing the learning of new classes with few-shot. CEC \cite{zhang2021few} introduces a graph attention model, combined with incremental learning. They get promising results by the algorithm updating only the parameters of the classification head during the incremental learning process. In \cite{10018890}, Ji \textit{et al.} proposes a Memorizing Complementation Network (MCNet) to ensemble multiple models that complement the different memorized knowledge with each other for novel classes.


     \textbf{Semi-supervised Learning} SSL aims to enhance the performance of a model by utilizing readily available unlabeled data in conjunction with a small amount of labeled data \cite{sohn2020fixmatch}, \cite{bachman2014learning}, \cite{berthelot2019mixmatch}, \cite{xie2020unsupervised}, \cite{lee2013pseudo}. SSL is typically implemented in two main approaches: consistency regularization \cite{sohn2020fixmatch}, \cite{bachman2014learning}, \cite{berthelot2019mixmatch}, \cite{xie2020unsupervised} and pseudo-labeling \cite{lee2013pseudo}, \cite{berthelot2019mixmatch} (also known as self-training). The fundamental concept behind consistency regularization is that when two inputs exhibit similarity in the input space, their outputs should also display similarity. This encourages the model to maintain consistency within the input space. On the other hand, pseudo-labeling involves using a model trained on labeled data to predict unlabeled data and assign pseudo-labels. Subsequently, the pseudo-labeled data is employed to augment the original dataset, contributing to the model's updates.

    \begin{figure*}[htb]
    \centering
    \includegraphics[height=8.6cm,width=\linewidth]{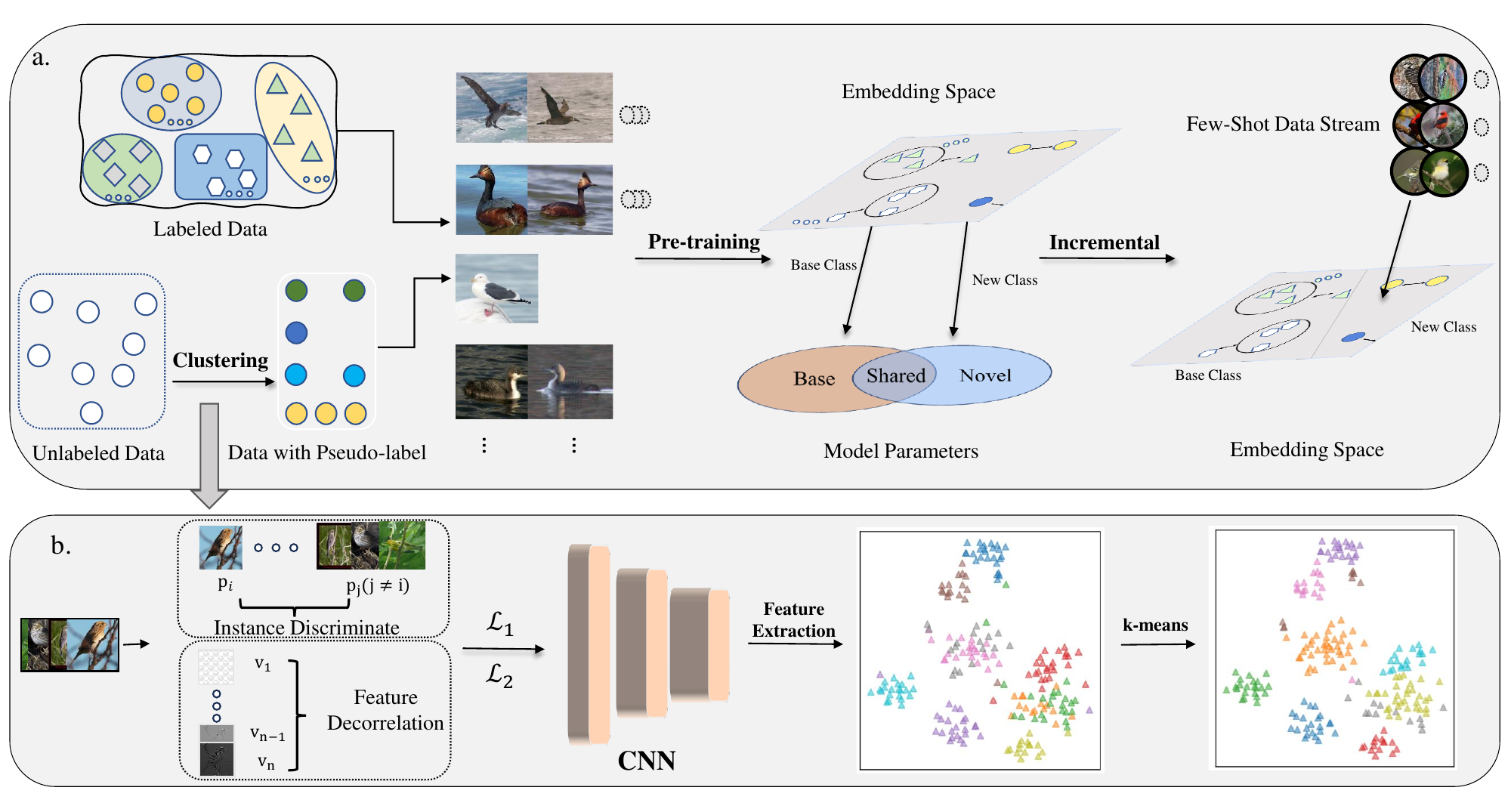}
    \caption{Overall Framework of LwPK. a) Pipeline of LwPK. In scenarios where the volume of labeled data surpasses that of unlabeled data. b) Detailed description of the clustering module. $p_i$ denotes the $i$-th image in the unlabeled dataset, $v_i$ denotes the $i$-th dimension of the feature. $\mathcal{L}_1$ and $\mathcal{L}_2$ are the two loss functions used to representation learning.}
    \label{fig:lc2}
    \end{figure*}

\section{Methods}
This section presents the LwPK algorithm designed for FSCIL, as shown in Figure \ref{fig:lc2}. LwPK comprises two crucial phases: a pre-training phase utilizing pseudo-labeled data obtained through deep clustering in conjunction with the base class data, and a phase dedicated to incremental learning on few-shot data. In the following, we will introduce the problem settings, method details, and related theoretical analysis.

\subsection{Problem Description}

In the few-shot class incremental learning task, we operate within a data stream \(\mathcal{D} = \{\mathcal{D}_0, \mathcal{D}_1, \ldots, \mathcal{D}_{n-1}, \mathcal{D}_n\}\). $n$ denotes incremental sessions. \(\mathcal{D}_0 = \left\{x_i, y_i\right\}^{B \times M}\) represents the base class dataset containing a substantial amount of labeled data. This initial dataset includes \(B\) classes with \(M\) instances per class and is utilized for training in the initial phase. Subsequently, data streams \(\mathcal{D}_s = \{x_i, y_i\}^{N \times K}\) \((1 \leq s \leq n)\) unfold sequentially, where each \(\mathcal{D}_s\) presents an \(N\)-way \(K\)-shot task, with \(B \gg N\) and \(M \gg K\). Incremental learning mainly addresses the issue of catastrophic forgetting, demanding the preservation of old knowledge memory when learning a new task upon the arrival of each data stream \(\mathcal{D}_s\). Therefore, after each model update, we take the accuracy of all the experienced categories as the main criterion to measure the model performance.

To construct LwPK, we aggregate unlabeled data from prospective classes to form \(\mathcal{D}_u = \{x_i \}^{n \times N \times U}\). Here, \(U\) represents the number of unlabeled data sheets selected for each class, and it's ensured that \(\mathcal{D}_u\) and \(\mathcal{D}_i\) do not exhibit any overlap. Leveraging clustering algorithm \(\mathcal{\phi}(\cdot)\), all unlabeled data in \(\mathcal{D}_u\) is organized into \(n \times N\) clusters, and each cluster is assigned a pseudo-label \(\mathcal{P}_i\). Consequently, we transform the original unlabeled dataset \(\mathcal{D}_u\) into a new dataset \(\mathcal{D}_p = \{x_i, p_i\}^{n \times N \times U}\), where the range of \(p_i\) values is maintained consistent with the label range of \(\mathcal{D}_i\). We combine \(\mathcal{D}_p\) with \(\mathcal{D}_0\) to obtain the joint dataset and simultaneously train both the feature extraction part \(\mathcal{F}\) and the classification head \(\mathcal{H}\) of the model.

\subsection{Prior Knowledge from Incremental Classes}
Assigning high-quality pseudo-labels to unlabeled data is a crucial challenge to obtain a supervisory signal. Currently, there are commercially available methods for few-shot multi-classification problems \cite{zhang2020deepemd,vinyals2016matching,snell2017prototypical} that have achieved good results. The problem we face is severer because we have no label information to serve as a basis for guiding network learning. Commonly employed self-supervised deep learning algorithms place higher demands on the dataset \cite{he2020momentum}, \cite{chen2020simple}. 

To tackle the challenge, we employ representation learning in conjunction with clustering (RLCC). Representation learning seeks to acquire meaningful representations or features of the input data \cite{bengio2013representation}, laying a crucial foundation for subsequent tasks. Our objective is to assign pseudo-labels to the data through clustering, emphasizing the desire for learned features to exhibit clustering-friendly characteristics. Inspired by the approach presented in \cite{tao2021clustering}, we incorporate instance discrimination \cite{wu2018unsupervised} and feature decorrelation to facilitate the learning of feature-friendly representations. Our ultimate goal is to obtain a feature representation model $f_r$ suitable for clustering.

In instance discrimination, each instance represents a category, acting as a positive sample for itself and a negative sample for other data points. For a given set of data \(\{x_1, x_2, \ldots, x_n\}\), we represent it with a set of feature vectors \(\{\mathbf{v}_1, \mathbf{v}_2, \ldots, \mathbf{v}_n\}\). Considering the similarity between images, the feature vector $\mathbf{v}$ of the image $\mathbf{x}$ can be used directly as weights. The probability that a data vector $\mathbf{v}$ is assigned to the $i$-class can be approximated as:

\begin{equation}
P(i\mid \mathbf{v}) = \frac{\exp(\mathbf{v_i^T} \mathbf{v}/ t_1)}{\sum_{j=1}^n \exp(\mathbf{v_j^T} \mathbf{v}/ t_1)}
\end{equation}
where $t_1$ is the temperature parameter. Maximizing the joint probability of $n$ classes of data $\prod_{i=1}^n P\left(i \mid f_r \left(x_i\right)\right)$ serves as the optimization objective. The objective function can be defined as:
\begin{equation}
\mathcal{L}_1=-\sum_{i=1}^n \log P\left(i \mid f_r\left(x_i\right)\right)
\end{equation}

Minimizing the loss function preserves the similarity between image instances and provides a more friendly representation for clustering \cite{wu2018unsupervised}.

The purpose of feature decorrelation is to construct independent features, which are defined as:
\begin{equation}
Q(i \mid \boldsymbol{f})=\frac{\exp \left(\boldsymbol{f}_i^T \boldsymbol{f} / t_2\right)}{\sum_{j=1}^d \exp \left(\boldsymbol{f}_j^T \boldsymbol{f} / t_2\right)}
\end{equation}
where $\boldsymbol{f}$ denotes a set of feature vectors and  $\boldsymbol{f}_i$ denotes the i-th dimension feature of the vector. $t_2$ is the temperature parameter. $Q(i\mid\boldsymbol{f})$ is used to compute the correlation of a feature vector with itself and the dissimilarity with other features. Similarly, the objective function can be defined as:
\begin{equation}
\mathcal{L}_2=-\sum_{i=1}^n \log Q(i \mid \boldsymbol{f})
\end{equation}

In contrast to the straightforward form of feature orthogonality, the softmax form of orthogonalization is evidently more lenient, yet it still can guide features toward optimization in the direction of orthogonality \cite{tao2021clustering}.

Combining the above two points, the final objective function is:
\begin{equation}
\mathcal{L}_u= \mathcal{L}_1 + \beta \mathcal{L}_2
\end{equation}
$\beta$ is used to balance the two loss functions.

\subsection{Training with Prior Knowledge}
\subsubsection{Hybrid embedding space due to joint training}
The challenge of new class underfitting and catastrophic forgetting in FSCIL often stems from a single model adaptation to either new class or base class data \cite{zou2022margin}. Striking a balance between model fitting and generalization proves challenging. In FSCIL, when the model is trained on base class data, it tends to adapt to the patterns of that data strongly. To mitigate this, we can introduce regularization to prevent overfitting on the base class data, improving generalization to some extent. However, this penalty-like approach makes it difficult to explicitly specify the model to optimize towards adapting to incremental data. Similarly, during the incremental process, the model needs to adjust itself to adapt to new data patterns. The overfitting problem arising from a small data volume can significantly impact model performance.

We construct a hybrid embedding space by combining base class data with pseudo-labeled data pairs, as illustrated in Figure \ref{fig:lc2}. This hybrid embedding space allocates separate regions for both base class and new class data. During the incremental stage, adjusting the model parameters becomes less pronounced, thereby avoiding space crowding issues caused by the model adapting to a specific mode. We introduce reconciliation weights $\omega$ to constrain the learning direction of the model. We take the cross-entropy loss with weights $\omega$ as the objective function:

\begin{equation}
\mathcal{L(\omega)}=-\frac{\omega}{N} \sum_{j=1}^M\sum_{i=1}^N \ p_{ij} \log q_{ij}
\end{equation}
where $p_{ij}$ is the ground truth of the $i$ sample, $q_{ij}$ is the score obtained from the model output $logits$ after the Softmax function. $N$ and $M$ represent the batch sizes and number of classes, respectively.

\subsubsection{Incremental learning with Prior Knowledge}
Owing to the inherent limitations of the data, applying conventional learning approaches to few-shot learning can lead to challenges in extracting the feature information present in the data or give rise to severe overfitting issues. ProtoNet \cite{snell2017prototypical} obtains a relatively general feature extractor by training on base class data, and uses it to obtain the average embedding of the subsequent few-shot data, defined as: 
\begin{equation}
\mathcal{E}_i= \frac{1}{|\mathcal{D}_i|} \sum_{i \in k} \mathcal{F}(\mathcal{D}[\mathbbm{1}[y=i]]);
\label{eq:em}
\end{equation}
where $\mathcal{F}()$ is the  feature extractor, and $x_i$ represents the data from the same class, $\mathcal{E}_i$ denotes the obtained average embedding of $i$-class.
We consider such an average embedding $\mathcal{E}_m$ as the prototype for the corresponding class data and use the prototype to update the parameters in the fully connected layer $W_b = \left[W_i:W_{i+1}\right] $ whenever new data comes in. 

\subsubsection{Incremental Inference}
In the preceding section, we substituted the parameters of the fully connected layer with the average embedding as a prototype. During the inference phase, it is essential to leverage these parameters as the foundation for classification. A conventional approach for this task is to employ cosine similarity $S(\mathbf{x})=\left(\frac{W}{\|W\|_2}\right)^{\top}\left(\frac{\mathcal{F}(x)}{\|\mathcal{F}(x)\|_2}\right)$.
We first extract the embedding of the input image using the feature extractor $\mathcal{F}()$ and then determine which class the input embedding should belong to by examining the cosine distance between the input embedding and the prototype. The probability $p$ that the input data $x$ belongs to the $i$-class can be calculated as:
\begin{equation}
P(i\mid x) = \frac{\exp(\mathbf{W_i^T} \mathbf{\mathcal{F}(x)})}{\sum_{\mathbf{W_i}\in\mathbf{W}} \exp(\mathbf{W_i^T} \mathbf{\mathcal{F}(x)})}
\end{equation}

The pseudo-code of LwPK is represented in Alg. \ref{alg:alg1}.

\begin{algorithm}[t]
\setstretch{1.2}
\algsetup{linenosize=\small} \small
\caption{\small{Pseudo code of LwPK}}
\label{alg:alg1}
\begin{algorithmic}
\STATE {\textbf{Input}: Labeled base data: \(\mathcal{D}_0\), unlabeled extra incremental data: \(\mathcal{D}_u\);}
\STATE {\textbf{Output}: \(\mathcal{F}\), \(\mathcal{H}\)}.
\STATE  1: Assign pseudo-labels: $ p_i = \Phi\left(x_i \right), x_i \in \mathcal{D}_u $;
\STATE  2: Blend the data to get a federated dataset: ${(x_i, y_i)}^n_{i=1}$;
\STATE \textbf{for} \underline{~} in range (epochs) \textbf{do}
\STATE \hspace{0.25cm} 3: Select mini-batch of training data randomly: ${(x_i, y_i)}^n_{i=1}$;
\STATE \hspace{0.25cm} 4: Design loss weights for additional data: $w_i$;
\STATE \hspace{0.25cm} 5: Calculate total forecasting loss: \(\mathcal{L}\);
\STATE \hspace{0.25cm} 6: Update the model  \(\mathcal{F}\), \(\mathcal{H}\);
\STATE 7: Get the average class data embedding: by Equation \ref{eq:em};
\STATE 8: Replace the FC layer with average embedding: $ \mathcal{H} \gets \mathcal{E}_m$.

\end{algorithmic}
\label{alg1}
\end{algorithm}

\subsection{Theoretical Analysis}
In \cite{germain2009pac},  Germain \textit{et al} introduced the concept of partitioning a portion of the training dataset to establish prior information. They further elaborated on how to optimally divide the dataset to achieve the highest performance upper bound using Equation \ref{eq2}. Our approach bears a resemblance to this concept, as both methods employ data to establish prior information for subsequent learning phases. The distinction lies in our decision not to split the subsequent dataset used for training since this operation is not compatible with incremental learning.

\begin{equation}
\begin{aligned}
& C m R\left(G_{Q_{\mathbf{w}}}\right)+\mathrm{KL}\left(Q_{\mathbf{w}} \| P_{\mathbf{w}_p}\right)= \\
& \quad C \sum_{i=1}^m \Phi\left(\frac{y_i \mathbf{w} \cdot \boldsymbol{\phi}\left(\mathbf{x}_i\right)}{\left\|\boldsymbol{\phi}\left(\mathbf{x}_i\right)\right\|}\right)+\frac{1}{2}\left\|\mathbf{w}-\mathbf{w}_p\right\|^2
\label{eq2}
\end{aligned}
\end{equation}

$C m R\left(\cdot \right)$ is the empirical risk of the decision, and the presence of $KL\left(\cdot \right)$ is done as a regularization. They point out that the absence of prior knowledge $w_p$ may lead to suboptimal solutions of the model.

We incorporate additional data as prior knowledge to construct pre-trained models. The influence of this knowledge on the incremental task can be examined from two perspectives: its impact on the base class and its impact on the new class. Specifically, we can regard this prior knowledge as the fundamental encoding in the feature space that influences the model's decision-making during prediction. With the integration of extra data, our training objective evolves to optimize the objective function
\begin{equation}
\begin{aligned}
& \min \frac{1}{n} \sum_{i=1}^n \mathcal{L}\left(\mathcal{F}_\theta\left(x_i\right), y_i\right)\\
& = \min \left(\frac{1}{n_B} \sum_{i=1}^{n_B} \mathcal{L}\left(\mathcal{F}_\theta\left(x_i\right), y_i\right) + \frac{\lambda}{n_E} \sum_{i=1}^{n_E} \mathcal{L}\left(\mathcal{F}_\theta\left(x_i\right), p_i\right)\right)
\label{eq3}
\end{aligned}
\end{equation}
where $n_B$ is the number of base class data $x_B$, $n_E$ is the number of additional data $x_E$, and $\lambda$ is the weight added for the additional data. $\mathcal{F}_\theta$ is a function that maps inputs to outputs. $p_i$ is the pseudo-label provided by the clustering algorithm for data $x_i\in x_E$. 

By optimizing the objective function, we obtain the parameters of the feature representation model $\theta$. We assume that $\theta$ can be composed of $\theta_b$ and $\theta_e$. $\theta_b$ is the parameter controlled by $x_B$ and $\theta_e$ is the parameter controlled by $x_E$. When making inferences on the base class data, we define the risk on the base class as
\begin{equation}
\begin{aligned}
\epsilon_B = \underset{b \in \mathcal{B}}{\mathbb{E}}\left[y_b \neq h\left(z_b\right) \mid z_b = \mathcal{R}\left(f_{\theta}\left(x_b\right)\right)\right]
\label{eq4} 
\end{aligned}
\end{equation}
where $R(\cdot)$ is the feature extraction function, $f(\cdot)$ is the parameter mapping function, $h(\cdot)$ is classification function, and $z_b$ is the feature representation of $x_b$. 

Due to the presence of additional classes, the performance is affected by the inter-class distance. This distance is primarily influenced by two factors: the characteristics of the image itself and the performance of the model fitted to the data. In the domain migration scenario, the goal is to generalize the model trained on the source domain to the target domain. The main factor that affects the model performance on the target domain is the difference in the properties of the two domains themselves, even though the model itself remains unchanged \cite{ben2006analysis}. In our task, the two training modes of adding incremental class data for joint training with the base class and pre-training followed by fine-tuning are different, and the final models obtained differ, so we need to consider both the model fitting data. Assume that the model parameters obtained by training on data consisting entirely of $x_B$ and entirely of $x_E$ are $\theta_B$ and $\theta_E$, respectively. We define
\begin{equation}
\begin{aligned}
\lvert f_{\theta}\left(x_b \right)-f_{\theta_B}\left(x_b \right) \rvert = m(\theta-\theta_B) = d_\theta\left(x_B, x_E \right) + \xi 
\label{eq5}
\end{aligned}
\end{equation}
where $m$ is a function that maps the difference of parameters. $d(\cdot)$ represents the distance between classes, we simplify the operation by using the distance of the average feature distribution as the inter-class distance. $\xi$ represents the bias caused by the characteristics of the data itself. In fact, the element inside $\lvert \cdot \rvert > 0$, $\lvert \cdot \rvert$ can be taken off directly.

In the first training mode,
\begin{equation}
    \begin{aligned}
        &\theta = (1-\alpha)\theta_B+\alpha\theta_E\\
        &\theta - \theta_B = \alpha(\theta_E - \theta_B)
    \end{aligned}
\end{equation}
where $\alpha$ is the impact factor of the additional data on the model, $\alpha \sim \frac{n_E}{n_B},\ 0<\alpha<1$.

In the second training mode, the model parameters turn into
\begin{equation}
    \begin{aligned}
        &\theta^{\prime} = (1-\beta)\theta_B+\beta\theta_E\\
        &\theta^{\prime} - \theta_B = \beta(\theta_E - \theta_B)
    \end{aligned}
\end{equation}
where $\beta$ is related to the degree of fine-tuning. 

At this point, the $z_b$ becomes
\begin{equation}
\begin{aligned}
z_b=\mathcal{R}\left(f_{\theta_B}\left(x_b\right)+m(\theta-\theta_B) \right) 
\label{eq7}
\end{aligned}
\end{equation}

As the amount of additional data we incorporate is small, $n_E \ll n_B$, this creates a long-tail problem where the model tends to focus more on the base class data. Therefore, $\alpha$ is relatively closer to 0. Due to the inevitable inter-class distance between the base class and the extra class, the model needs to adjust itself to fit the new data, the effect of $\beta$ deepens further. There will be $\beta > \alpha$, leading to
\begin{equation}
\begin{aligned}
\epsilon_B < \epsilon_B^{\prime} \leq \underset{b \in \mathcal{B}}{\mathbb{E}}\left[y_b \neq h\left(z_b\right) \mid z_b = f_{\theta_E}\left(x_b\right)\right)]
\label{eq9}
\end{aligned}
\end{equation}

Therefore, adding additional knowledge in the pre-training phase reduces the parameter adjustment during the incremental process and enhances the memory capability of the model. 

In addition, since we address the few-shot data problem, the conventional training approach can easily overfit these data. We use feature extraction with the calculation of cosine similarity to determine the attribution of the data, and the risk on the new class data, if no additional data is added, can be defined as
\begin{equation}
\begin{aligned}
\epsilon_N &= \underset{n \in \mathcal{N}}{\mathbb{E}}\left[y_n \neq h\left(z_n\right) \mid z_n = \mathcal{R}(\mathcal{f}_{\theta_B}(x_n))\right] \\
&= \underset{n \in \mathcal{N}}{\mathbb{E}}\left[y_n \neq h\left(z_n\right) \mid z_n = \mathcal{R}(f_{\theta_{B}}(x_b) + d_\theta(x_B, x_N) + \xi)\right]
\label{eq10}
\end{aligned}
\end{equation}

This does not diminish the impact of inter-class distance and does not improve the model's ability to extract features from the new class data. We set the influence factor of $\theta_e$ on $f$ as $\varepsilon, \varepsilon \sim \frac{n_E}{n_B},\ 0<\varepsilon<1$, the feature representation $z_n$ can turn into 
\begin{equation}
\begin{aligned}
z_n^{\prime}=&\mathcal{R}\left( (1-\varepsilon) f_{\theta_b}\left(x_n\right)+\varepsilon f_{\theta_e} \left(x_n \right) \right)\\
=&\mathcal{R} ( (1-\varepsilon) \lvert f_{\theta_b} \left( x_b \right)+ d_\theta\left(x_B, x_N \right)+\xi \rvert \\
& + \varepsilon \lvert f_{\theta_e} \left(x_e \right)+ d_\theta\left(x_E, x_N \right)+\xi \rvert) 
\label{eq11} 
\end{aligned}
\end{equation}
    
When we use the inter-class distance to measure the risk of model on the new class, it is worth noting that here we need to consider the accuracy of the clustering algorithm. 

Suppose there exists an ideal representation model that can aggregate data belonging to the same class and separate data belonging to different classes on the ideal feature space $\mathcal{S}_i$. Let the average intra-class distance $D_{in}$ and the average inter-class distance $D_{out}$, $D_{in} \textless D_{out}$. In the actual absence of an ideal representation model, we cannot map the data ideally. The mapping of the data on the feature space deviates from $\mathcal{S}_i$, thus leading to incorrect clustering. 

In the subsequent training, the model adjusts its own parameters to fit the data according to the purpose of optimizing the loss function. Due to the error information generated by the previous clustering, the model cannot map the data ideally, and the feature space $\mathcal{S}_r$ still deviates from $\mathcal{S}_i$. Assuming that the clustering accuracy is $\mathcal{A}$, we can define the offset distance of the new data $x_n$ in $\mathcal{S}_i$ and $\mathcal{S}_r$ as $D = \mathcal{A}D_{in}+(1-\mathcal{A})D_{out}$. In the feature space obtained from the random model mapping, the different classes cannot be truly separated from each other, and at this point, the average inter-class distance is assumed to be $D_{rand}$, $D_{in} < D_{rand}< D_{out}$. When the model is trained on the base class data only, we consider that the model does not do anything with the new data and its performance on the new class approximates the performance of the random model, $d_\theta\left(x_B, x_N \right) \approx D_{rand}$. Similarly, $d_\theta\left(x_E, x_N \right) \approx D$. 

Therefore, there exists an accuracy $\mathcal{A}_0$ when $\mathcal{A} > \mathcal{A}_0$, $D<D_{rand}$.
 At this time,
\begin{equation}
\begin{aligned}
d\left(x_E, x_N \right) &< d\left(x_B, x_N \right)\\
\epsilon_N^{\prime} &< \epsilon_N.
\label{eq12} 
\end{aligned}
\end{equation}

In summary, by incorporating prior knowledge during the pre-training phase, we can preserve the model memory and achieve improved performance on the new class with a performance clustering algorithm.

\section{Experiments}
In this section, we carry out experiments on several benchmark datasets for FSCIL, namely CIFAR100, CUB200, and {\it mini}ImageNet. We compare our experimental outcomes with the baseline and current state-of-the-art (SOTA) to demonstrate the feasibility of LwPK.

\subsection{Experiment Details}

\begin{table*}[htb]
\caption{INTRODUCTION OF THREE BENCHMARK DATASETS.}
\label{tab:1}
\centering
\begin{tabular}{c|c|c|c|c|c|c|c}
\hline
\multirow{2}{*}{\centering Datasets} & \multirow{2}{*}{\centering Total Classes} & \multicolumn{2}{c|}{\centering Session 0} & \multicolumn{3}{c|}{Session \it{i} (\it{i}\textgreater0)}  & \multirow{2}{*}{\centering Img Size} \\

\cline{3-7}
& & Classes & Samples & Classes & Samples & Session Num
 \\
\hline
CUB200  & 200 & 100 & 30 & 10 & 5 & 10 &224$\times$224 \\
CIFAR100  & 100 & 60 & 500 & 5 & 5 & 8 & 32$\times$32 \\
{\it mini}ImageNet & 100 & 60 & 500 & 5 & 5 & 8 & 84$\times$84 \\
\hline
\end{tabular}
\end{table*}

\subsubsection{Data Configuration}
    The foundational configurations of the three datasets, CIFAR100, CUB200, and {\it mini}ImageNet, are illustrated in Table \ref{tab:1}. The delineation criteria align with \cite{tao2020few}.
\subsubsection{Training Configuration}
    All models are implemented in PyTorch, and the model selection aligns with \cite{tao2020few}. Specifically, for the CIFAR100 dataset, we employed ResNet20 \cite{he2016deep}, while for the CUB200 and miniImageNet datasets, ResNet18 was utilized. We use SGD with momentum for optimization. Two learning rates, 0.1 and 5e-3, are employed for training from scratch and fine-tuning, respectively.
    In the clustering task, we use ResNet18 as a feature extractor to obtain the feature distribution of all unlabeled data, and then perform the clustering operation based on these features. Specifically for the CUB200 dataset, we utilized the same model setup as in incremental learning and directly employed the pre-trained ResNet18 for feature extraction \cite{tao2020few}.
\subsubsection{Evaluation Tools}
    Following \cite{zhou2022forward}, we choose top-1 accuracy as our evaluation metric to test the recognition accuracy of the model for all the emerged category data after each session. We also test the performance degradation (PD) of the model from session 0 to the end as a criterion to judge the memory ability of the model. In addition, we selected the accuracy metrics for evaluation, including the accuracy of the first session ($Acc_f$), the accuracy of the last session ($Acc_l$), and the average accuracy across all stages ($Acc_{avg}$).

\subsection{Benchmark Comparison}

\begin{table*}[tb]         
\small 
\renewcommand\arraystretch{1.3}
\caption{TEST ACCURACY OF EACH INCREMENTAL SESSION ON CUB200 DATASET.}
\label{tab11}
\centering
\resizebox{\linewidth}{!}{\begin{tabular}{c c c c c c c c c c c c c c c c c c}
\toprule
\multicolumn{1}{c}{\multirow{2}{*}{Task}} & \multicolumn{1}{c}{\multirow{2}{*}{Method}} & \multicolumn{11}{c}{Accuracy in each session ($\%$)}&\multirow{2}{*}{PD $\downarrow$} &\multirow{2}{*}{$Acc_{avg}$}\\
\cline{3-13}
& &0 & 1 & 2 & 3 & 4 & 5 & 6 & 7 & 8 & 9 & 10 \\ 
\hline
\thead{\normalsize \multirow{5}{*}[-3.2ex]{CIL}} & \thead{Finetune} &\thead{68.68} & \thead{43.70} & \thead{25.05} & \thead{17.72} & \thead{18.08} & \thead{16.95} & \thead{15.10} &\thead{10.06} & \thead{8.93} & \thead{8.93} & \thead{8.47} & \thead{60.21} & \thead{21.97} \\

&\thead{Joint} &\thead{78.93} & \thead{74.78} & \thead{72.33} & \thead{68.01} & \thead{67.71} & \thead{64.81} & \thead{64.26} &\thead{63.74} & \thead{62.93} & \thead{62.70} & \thead{62.11} & \thead{16.82} & \thead{67.48}\\

&\thead{iCaRL \cite{rebuffi2017icarl}} &\thead{68.68} & \thead{52.65} & \thead{48.61} & \thead{44.16} & \thead{36.62} & \thead{29.52} & \thead{27.83} &\thead{26.26} & \thead{24.01} & \thead{23.89} & \thead{21.16} & \thead{47.52} & \thead{36.67}\\

&\thead{EEIL \cite{castro2018end}} &\thead{68.68} & \thead{53.63} & \thead{47.91} & \thead{44.20} & \thead{36.30} & \thead{27.46} & \thead{25.93} &\thead{24.70} & \thead{23.95} & \thead{24.13} & \thead{22.11} & \thead{46.57} & \thead{36.27}\\

&\thead{Rebalancing \cite{hou2019learning}} &\thead{68.68} & \thead{57.12} & \thead{44.21} & \thead{28.78} & \thead{26.71} & \thead{25.66} & \thead{24.62} &\thead{21.52} & \thead{20.12} & \thead{20.06} & \thead{19.87} & \thead{48.81} & \thead{32.49}\\
\hline
\thead{\normalsize \multirow{4}{*}[-2.4ex]{FSCIL}}
&\thead{TOPIC \cite{tao2020few}} &\thead{68.68} & \thead{62.49} & \thead{54.81} & \thead{49.99} & \thead{45.25} & \thead{41.40} & \thead{38.35} &\thead{35.36} & \thead{32.22} & \thead{28.31} & \thead{26.26} & \thead{42.40} & \thead{43.92}\\

&\thead{Decoupled-DeepEMD \cite{zhang2020deepemd}} &\thead{75.35} & \thead{70.69} & \thead{66.68} & \thead{62.34} & \thead{59.76} & \thead{56.54} & \thead{54.61} &\thead{52.52} & \thead{50.73} & \thead{49.20} & \thead{47.60} & \thead{27.75} & \thead{58.73}\\

&\thead{CEC \cite{zhang2021few}} &\thead{75.85} & \thead{71.94}  & \thead{68.50} & \thead{63.50} & \thead{62.43} & \thead{58.27} &\thead{57.73} & \thead{55.81} & \thead{54.83} & \thead{53.52} & \thead{52.28} & \thead{23.57} & \thead{61.33}\\

&\thead{FACT \cite{zhou2022forward}} &\thead{75.90} & \thead{73.23} & \thead{70.84} & \thead{66.13} & \thead{65.56} & \thead{62.15} & \thead{61.74} &\thead{59.83} & \thead{58.41} & \thead{57.89} & \thead{56.94} & \thead{18.96} & \thead{64.42}\\

&\thead{MCNet \cite{10018890}} &\thead{77.57} & \thead{73.96} & \thead{70.47} & \thead{65.81} & \thead{66.16} & \thead{63.81} & \thead{62.09} &\thead{61.82} & \thead{60.41} & \thead{60.09} & \thead{59.08} & \thead{18.49} & \thead{65.57}\\
\hline
\thead{\normalsize \multirow{3}{*}[-1.6ex]{Semi-FSCIL}}
&\thead{SS-NCM-CNN \cite{cui2021semi}} &\thead{69.89} & \thead{64.87} & \thead{59.82} & \thead{55.14} & \thead{52.48} & \thead{49.60} & \thead{47.87} &\thead{45.10} & \thead{40.47} & \thead{38.10} & \thead{35.25} & \thead{34.64} & \thead{50.78}\\

&\thead{Us-KD \cite{cui2022uncertainty}} &\thead{74.69} & \thead{71.71} & \thead{69.04} & \thead{65.08} & \thead{63.60} & \thead{60.96} & \thead{59.06} &\thead{58.68} & \thead{57.01} & \thead{56.41} & \thead{55.54} & \thead{19.15} & \thead{62.89}\\

&\thead{UaD-ClE \cite{cui2023uncertainty}} &\thead{75.17} & \thead{73.27} & \thead{70.87} & \thead{67.14} & \thead{65.49} & \thead{63.66} & \thead{62.42} & \textbf{\thead{62.55}} & \textbf{ \thead{60.99}} & \textbf{\thead{60.48}} & \textbf{\thead{60.72}} & \textbf{\thead{14.45}} & \thead{65.70}\\
\hline
&\textbf{\thead{LwPK(Ours)}} & \textbf{\thead{78.30}} & \textbf{\thead{74.82}} & \textbf{\thead{71.90}} & \textbf{\thead{67.58}} & \textbf{\thead{66.83}} & \textbf{\thead{64.25}} & \textbf{\thead{62.92}} &\thead{61.59} & \thead{60.65} & \thead{59.65} & \thead{58.69} & \thead{19.61} & \bf{66.11}\\
\bottomrule
\label{tab:2}
\end{tabular}}
\end{table*} 

\begin{table*}[htbp]         
\small
\renewcommand\arraystretch{1.3}
\caption{TEST ACCURACY OF EACH INCREMENTAL SESSION ON CIFAR100 DATASET.}
\label{tab11}
\centering
\resizebox{\linewidth}{!}{
\begin{tabular}{c c c c c c c c c c c c c c c c}
\toprule
\multicolumn{1}{c}{\multirow{2}{*}{Task}} & \multicolumn{1}{c}{\multirow{2}{*}{Method}} & \multicolumn{9}{c}{Accuracy in each session ($\%$)}&\multirow{2}{*}{PD $\downarrow$} &\multirow{2}{*}{$Acc_{avg}$}\\
\cline{3-11}
& &0 & 1 & 2 & 3 & 4 & 5 & 6 & 7 & 8 \\ 
\hline




\thead{\normalsize \multirow{4}{*}[-2.4ex]{FSCIL}}
&\thead{TOPIC \cite{tao2020few}} &\thead{64.10} & \thead{55.88} & \thead{47.07} & \thead{45.16} & \thead{40.11} & \thead{36.38} & \thead{33.96} &\thead{31.35} & \thead{29.37} & \thead{34.73} & \thead{42.62}\\

&\thead{Decoupled-DeepEMD \cite{zhang2020deepemd}} &\thead{69.75} & \thead{65.06} & \thead{61.20} & \thead{57.21} & \thead{53.88} & \thead{51.40} & \thead{48.80} &\thead{46.84} & \thead{44.41} & \thead{25.34} & \thead{55.39}\\

&\thead{CEC \cite{zhang2021few}} &\thead{73.07} & \thead{68.88}  & \thead{65.26} & \thead{61.19} & \thead{58.09} & \thead{55.57} &\thead{53.22} & \thead{51.34} & \thead{49.14} & \thead{23.93} & \thead{59.53}\\

&\thead{FACT \cite{zhou2022forward}} &\thead{74.60} & \thead{72.09} & \thead{67.56} & \thead{63.52} & \thead{61.38} & \thead{58.36} & \thead{56.28} &\thead{54.24} & \thead{52.10}  & \thead{22.50} & \thead{62.24}\\

&\thead{MCNet \cite{10018890}} &\thead{77.57} & \thead{73.96} & \thead{70.47} & \thead{65.81} & \thead{66.16} & \thead{63.81} & \thead{62.09} &\thead{61.82} & \thead{60.41}  & \thead{18.49} & \thead{65.57}\\

\hline
\thead{\normalsize \multirow{4}{*}[-1.6ex]{Semi-FSCIL}}
&\thead{SS-NCM-CNN \cite{cui2021semi}} &\thead{64.13} & \thead{62.29} & \thead{61.31} & \thead{57.96} & \thead{54.26} & \thead{50.95} & \thead{49.02} &\thead{45.85} & \thead{44.59}  & \textbf{\thead{19.54}} & \thead{54.51}\\

&\thead{Us-KD \cite{cui2022uncertainty}} &\thead{76.85} & \thead{69.87} & \thead{65.46} & \thead{62.36} & \thead{59.86} & \thead{57.29} & \thead{55.22} &\thead{54.91} & \thead{54.42}  & \thead{22.43} & \thead{61.80}\\

&\thead{UaD-ClE \cite{cui2023uncertainty}} &\thead{75.55} & \thead{72.17} & \thead{68.57} & \thead{65.35} & \thead{62.80} & \thead{60.27} & \thead{59.12} &\thead{57.05} & \thead{54.50}  & \thead{21.05} & \thead{63.93}\\

\hline

&\textbf{\thead{LwPK(Ours)}} & \textbf{\thead{78.52}} & \textbf{\thead{73.09}} & \textbf{\thead{70.37}} & \textbf{\thead{66.15}} & \textbf{\thead{63.94}} & \textbf{\thead{61.69}} & \textbf{\thead{59.91}} & \textbf{\thead{58.00}} & \textbf{\thead{55.95}} & \thead{22.57} & \bf{\thead{65.29}} \\

\bottomrule
\label{tab:3}
\end{tabular}
}
\end{table*} 

\begin{table*}[htbp]         
\small
\renewcommand\arraystretch{1.3}
\caption{TEST ACCURACY OF EACH INCREMENTAL SESSION ON {\it mini}IMAGENET DATASET.}
\label{tab11}
\centering
\resizebox{\linewidth}{!}{
\begin{tabular}{c c c c c c c c c c c c c c c c c c}
\toprule
\multicolumn{1}{c}{\multirow{2}{*}{Task}} & \multicolumn{1}{c}{\multirow{2}{*}{Method}} & \multicolumn{9}{c}{Accuracy in each session ($\%$)}&\multirow{2}{*}{PD $\downarrow$} &\multirow{2}{*}{$Acc_{avg}$}\\
\cline{3-11}
& &0 & 1 & 2 & 3 & 4 & 5 & 6 & 7 & 8 \\ 





\hline
\thead{\normalsize \multirow{4}{*}[-1.6ex]{FSCIL}}
& \thead{TOPIC \cite{tao2020few}} &\thead{61.31} & \thead{50.09} & \thead{45.17} & \thead{41.16} & \thead{37.48} & \thead{35.52} & \thead{32.19} &\thead{29.46} & \thead{24.42} &  \thead{36.89} & \thead{39.64}\\

&\thead{Decoupled-DeepEMD \cite{zhang2020deepemd}} &\thead{69.77} & \thead{64.59} & \thead{60.21} & \thead{56.63} & \thead{53.16} & \thead{50.13} & \thead{47.79} &\thead{45.42} & \thead{24.35} & \thead{27.75} & \thead{54.57}\\

&\thead{CEC \cite{zhang2021few}} &\thead{72.00} & \thead{66.83}  & \thead{62.97} & \thead{59.43} & \thead{56.70} & \thead{53.73} &\thead{51.19} & \thead{49.24} & \thead{47.63} &  \thead{24.37} & \thead{57.75}\\

&\thead{FACT \cite{zhou2022forward}} &\thead{72.56} & \thead{69.63} & \thead{66.38} & \thead{62.77} & \thead{60.60} & \thead{57.33} & \thead{54.34} &\thead{52.16} & \thead{50.49} & \thead{22.07} & \thead{60.70}\\

&\thead{MCNet \cite{10018890}} &\thead{77.57} & \thead{73.96} & \thead{70.47} & \thead{65.81} & \thead{66.16} & \thead{63.81} & \thead{62.09} &\thead{61.82} & \thead{60.41}  & \thead{18.49} & \thead{65.57}\\

\hline
\thead{\normalsize \multirow{4}{*}[-1.6ex]{Semi-FSCIL}}
&\thead{SS-NCM-CNN \cite{cui2021semi}} &\thead{62.98} & \thead{60.88} & \thead{57.63} & \thead{52.80} & \thead{50.66} & \thead{48.28} & \thead{45.27} &\thead{41.65} & \thead{40.51}  & \thead{22.47} & \thead{51.26}\\

&\thead{Us-KD \cite{cui2022uncertainty}} &\thead{72.35} & \thead{67.22} & \thead{62.41} & \thead{59.85} & \thead{57.81} & \thead{55.52} & \thead{52.64} &\thead{50.86} & \thead{50.47}  & \thead{21.88} & \thead{58.79}\\

&\thead{UaD-ClE \cite{cui2023uncertainty}} &\thead{72.35} & \thead{66.91} & \thead{62.13} & \thead{59.89} & \thead{57.41} & \thead{55.52} & \thead{53.26} &\thead{51.46} & \thead{50.52}  & \thead{21.83} & \thead{58.82}\\
\hline
&\textbf{\thead{LwPK(Ours)}} & \textbf{\thead{75.72}} & \textbf{\thead{71.62}} & \textbf{\thead{68.11}} & \textbf{\thead{65.87}} & \textbf{\thead{63.59}} & \textbf{\thead{61.25}} & \textbf{\thead{59.24}} &\textbf{\thead{58.60}} & \textbf{\thead{57.84}} & \textbf{\thead{17.88}} & \bf{\thead{64.64}}\\

\bottomrule
\label{tab:4}
\end{tabular}
}
\end{table*} 

\begin{figure*}[t]
    \centering
    \includegraphics[height=5.1cm,width=\linewidth]{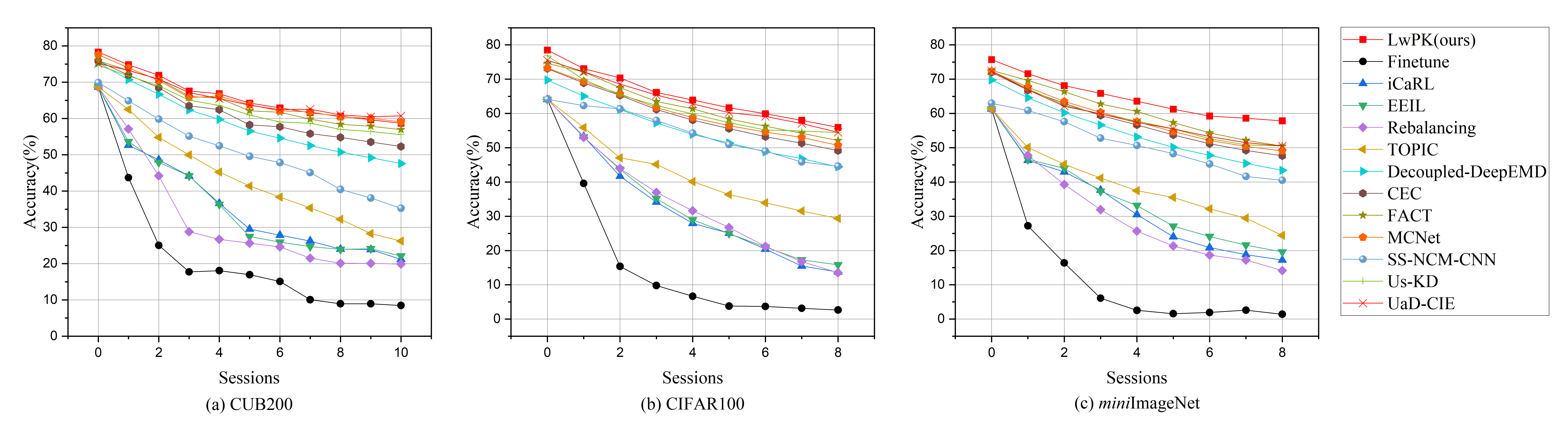}
    \caption{Test accuracy of each session on three datasets.}
    \label{fig:res}
\end{figure*}

In this section, we compare LwPK with the current mainstream and SOTA methods, including the traditional class incremental learning models iCaRL \cite{rebuffi2017icarl}, EEIL \cite{castro2018end}, Rebalancing \cite{hou2019learning}, as well as the FSCIL models TOPIC \cite{tao2020few}, Decoupled-DeepEMD \cite{zhang2020deepemd}, CEC \cite{zhang2021few} and FACT \cite{zhou2022forward} for few-shot. Additionally, we compare with several methodologies leveraging unlabeled data, including UaD-ClE \cite{cui2023uncertainty}, SS-NCM-CNN \cite{cui2021semi}, Us-KD \cite{cui2022uncertainty}. It is worth noting that these methods using unlabeled data are based on semi-supervised learning, which is different from LwPK. In the previous section, we described the relevant information about `prior' in our method, and similarly, we can divide these methods into `prior' and `prior less' categories. To more intuitively represent the performance of our method, we also give upper bounds (joint training) and lower bounds (finetune) for FSCIL tasks. All experimental data are shown in Figure \ref{fig:res} and Table \ref{tab:2}, \ref{tab:3}, \ref{tab:4}.

We can see that LwPK has better performance on few-shot tasks compared to traditional class incremental learning methods. Also, LwPK outperforms many FSCIL methods in most cases. FACT is a `prior' method that provides prior knowledge for subsequent data by predicting virtual instances. Although LwPK is a little more severe in terms of PD than FACT, the test accuracy on all sessions is higher. LwPK also has advantages over these semi-supervised learning methods that use unlabeled data of incremental classes.

It is normal that LwPK falls short of the upper bound represented by supervised training, and the problem may come from the clustering accuracy. 

\subsection{Ablation Study}
\begin{figure}[htbp]
    \centering
    \includegraphics[height=9.2cm,width=8.6cm]{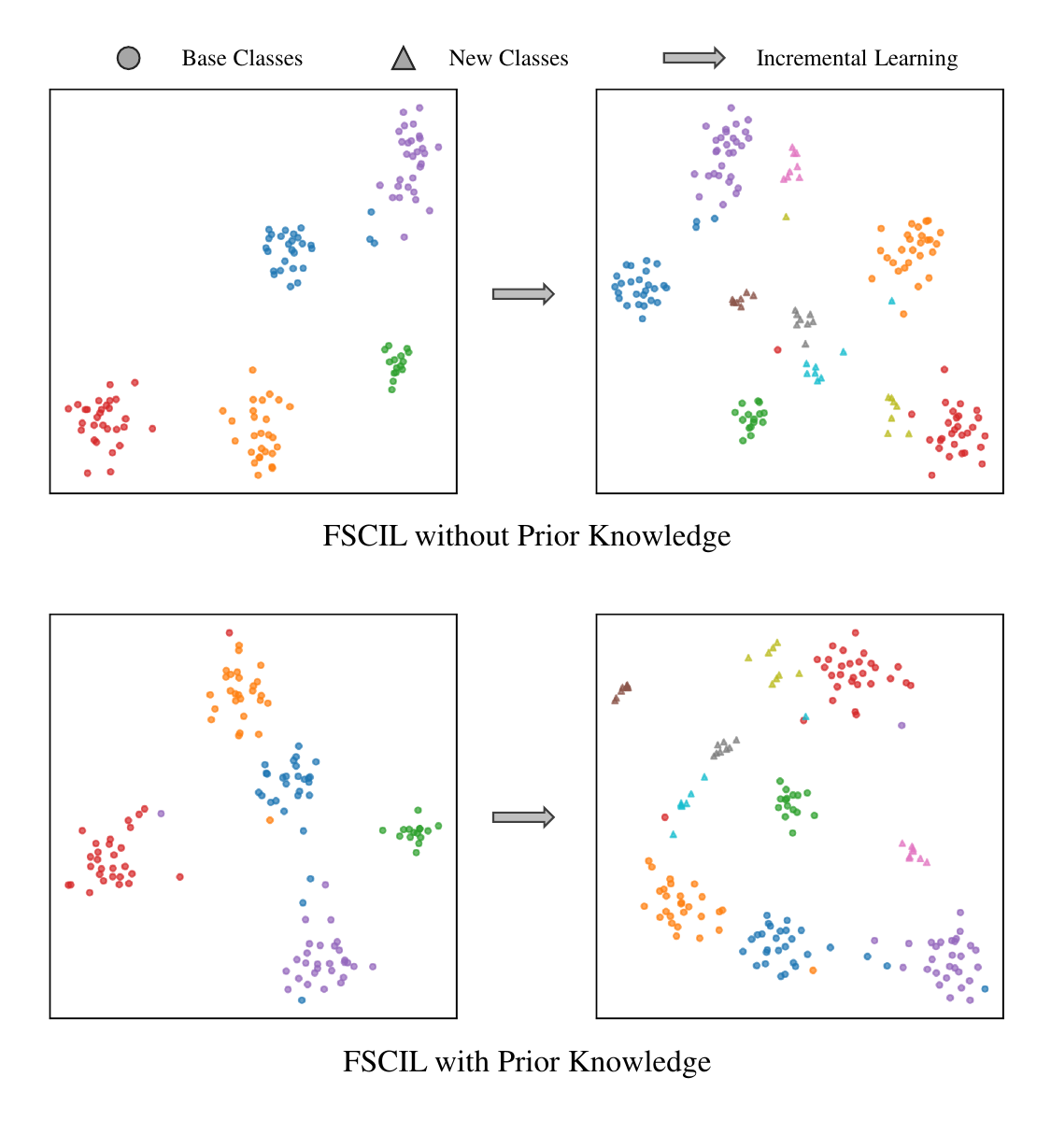}   \caption{t-SNE visualization plot, 5 base classes \& 5 new classes on CUB200.}
    \label{fig:res3}
    \end{figure}

    \textbf{Influence of Prior Knowledge} We leverage the embedding of prior knowledge within the pre-trained model to enhance performance specifically for the new class. To visually demonstrate the effectiveness of prior knowledge, the model performance differences before and after incorporating prior knowledge on both the base class data and the new class data are compared. The results are illustrated in Figure \ref{fig:res3}. We observe that the inclusion of prior knowledge improves the performance on new class data without adversely affecting the distribution of features from the base classes. Additionally, we present comprehensive results in Table \ref{tab:res3}, demonstrating the widespread enhancement of the model performance across all datasets.

    \begin{table}[!htbp] 
    \renewcommand\arraystretch{1.8}
    \normalsize
    \centering
    \caption{INFLUENCE OF PRIOR KNOWLEDGE (PK) ON ALL DATASETS.}
    \setlength{\tabcolsep}{4pt}{
    \label{tab:res3}
    \begin{tabular}{c c c c c}
    \bottomrule
    {Datasets} & PK & $Acc_f$ & $Acc_l$ & $Acc_{avg}$ \\
    \hline
    
    \thead{\normalsize \multirow{2}{*}[-1.6ex]{CIFAR100}} &\thead{\small \XSolid} &\thead{\normalsize 78.67} & \thead{\normalsize 51.94} & \thead{\normalsize 62.72}  \\
     & \thead{\small \Checkmark} &\thead{\normalsize 78.52} & \thead{\normalsize 55.95} & \thead{\normalsize 65.29} \\
     \hline
     
    \thead{\normalsize \multirow{2}{*}[-1.6ex]{{\it mini}ImageNet}}  &\thead{\small \XSolid} & \thead{\normalsize 72.62} & \thead{\normalsize 47.87}  & \thead{\normalsize 58.36} \\
     & \thead{\small \Checkmark} & \thead{\normalsize 75.72} & \thead{\normalsize 57.84}  & \thead{\normalsize 64.64} \\
     \hline
    
    \thead{\normalsize \multirow{2}{*}[-1.6ex]{CUB200}} & \thead{\small \XSolid} & \thead{\normalsize 78.85} & \thead{\normalsize 54.98} & \thead{\normalsize 63.69}\\
    
     & \thead{\small \Checkmark} & \thead{\normalsize 78.30} & \thead{\normalsize 58.69} & \thead{\normalsize 66.11}\\
    \toprule
    \end{tabular}}
    \end{table}

    \textbf{Influence of label mismatch} As clustering is employed to assign pseudo-labels to unlabeled data, the pseudo-labeling cannot be given in accordance with the original true-value labeling pattern, and to verify the impact of the labeling error present in it, we chose the limiting case of one sample per class, where each sample was given a label that may not match the labeling values of the subsequent incremental learning process, the result is shown in Table \ref{tab:res4}. The results indicate that label mismatch does not compromise the final experimental outcomes. This provides proof of the feasibility of the subsequent method.

    \begin{table}[!htbp]    
    \renewcommand\arraystretch{1.8}
    \normalsize
    \centering
    \caption{INFLUENCE OF LABEL MISMATCH ON ALL DATASETS. ``G" AND ``P" REPRESENT GROUND TRUTH AND PSEUDO-LABELS, RESPECTIVELY.}
    \setlength{\tabcolsep}{4pt}{
    \label{tab:res4}
    \begin{tabular}{c c c c c}
    \bottomrule
    {Datasets} & G/P & $Acc_f$ & $Acc_l$ & $Acc_{avg}$  \\
    \hline
    
    \thead{\normalsize \multirow{2}{*}[-1.6ex]{CIFAR100}} &\thead{\normalsize G} &\thead{\normalsize 78.82} & \thead{\normalsize 51.69} & \thead{\normalsize 62.89}  \\
    
     & \thead{\small P} &\thead{\normalsize 78.80} & \thead{\normalsize 51.78} & \thead{\normalsize 63.02} \\
     \hline
     
    \thead{\normalsize \multirow{2}{*}[-1.6ex]{{\it mini}ImageNet}}  &\thead{\normalsize G} & \thead{\normalsize 72.70} & \thead{\normalsize 47.49}  & \thead{\normalsize 57.64} \\
    
     & \thead{\normalsize P} & \thead{\normalsize 72.97} & \thead{\normalsize 48.06}  & \thead{\normalsize 58.47} \\
     \hline
    
    \thead{\normalsize \multirow{2}{*}[-1.6ex]{CUB200}}  &\thead{\normalsize G} & \thead{\normalsize 78.90} & \thead{\normalsize 57.79} & \thead{\normalsize 65.31}\\
    
     & \thead{\normalsize P} & \thead{\normalsize 78.90} & \thead{\normalsize 58.03} & \thead{\normalsize 65.82}\\
    \toprule
    \end{tabular}}
    \end{table}

    \begin{table}[!htbp]    
    \renewcommand\arraystretch{1.8}
    \normalsize
    \centering
    \caption{ARI OF DIFFERENT CLUSTERING ALGORITHMS FOR THREE DATASETS. THE VALUE OF ARI IS DISTRIBUTED AS [-1, 1], HIGHER REPRESENTS BETTER PERFORMANCE. R18$\textsuperscript{†}$ DENOTES THE USE OF THE PRE-TRAINED ResNet18.}
    \setlength{\tabcolsep}{4pt}{
    \label{tab:res6}
    \begin{tabular}{c c c c}
    \bottomrule
    {Method} & CIFAR100 &  {\it mini}ImageNet & CUB200  \\
    \hline
    \thead{\normalsize UniSiam} &\thead{\normalsize 0.04} & \thead{\normalsize 0.08} & \thead{\normalsize 0.02}  \\
    \thead{\normalsize ScatSimCLR} & \thead{\normalsize 0.05} & \thead{\normalsize 0.06}  & \thead{\normalsize 0.02} \\
    \thead{\normalsize R18$\textsuperscript{†}$} & \thead{\normalsize 0.12} & \thead{\normalsize 0.21}  & \thead{\normalsize 0.07} \\
    \hline
    \thead{\normalsize RLCC} & \thead{\normalsize 0.30} & \thead{\normalsize 0.38} & \thead{\normalsize -} \\
    \toprule
    \end{tabular}}
    \end{table} 
    
    \textbf{Accuracy of pseudo-label} In the previous discussion, we have mentioned the effect of pseudo-label quality on the final incremental learning. To obtain high-quality pseudo-labels, we tried many kinds of clustering algorithms, including  UniSiam \cite{lu2022self}, ScatSimCLR \cite{kinakh2021scatsimclr}. Due to possible label mismatches, we use the adjusted rand index (ARI) \cite{hubert1985comparing} to show the goodness of clustering. 

    We show the final clustering results in Table \ref{tab:res6}. From the experimental results, we can see that the existing unsupervised clustering algorithms are also difficult to perform in the face of few-shot multiclass tasks. LwPK can still achieve good performance in the incremental phase without precise clustering. Such results illustrate that pre-trained models can play an important role in incremental learning and that adding prior knowledge in the pre-training phase is helpful to enhance subsequent learning.

    \textbf{Impact of the amount of unlabeled data} The quantity of unlabeled data is a significant factor, directly influencing the construction of the pre-trained model. We conducted comparative experiments to illustrate the impact of data volume on the final results. For the CUB200 dataset, we opted for a more refined and randomized selection process. The outcomes are presented in Table \ref{tab:res7}. The table reveals that the quantity of selected data does not consistently follow the principle of "more is better." Instead, there exists a trade-off between valid and erroneous information. This phenomenon is primarily attributed to clustering accuracy issues, where incorrect assignment of pseudo-labels equates to introducing noise. Such noise can negatively impact the model. This problem will be slowed down with the improvement of clustering accuracy, and we will follow up with more research in this area.

\begin{table*}[bpht]
\small 
\renewcommand\arraystretch{1.3}
\caption{IMPACT OF THE AMOUNT OF UNLABELED DATA PER CLASS. UPC REPRESENTS THE NUMBER OF UNLABELED SAMPLES TAKEN FOR EACH CLASS.}
\label{tab11}
\centering
\resizebox{\linewidth}{!}{\begin{tabular}{c c c c c c c c c c c c c c c c c}
\toprule
\multicolumn{1}{c}{\multirow{2}{*}{Datasets}}  &\multirow{2}{*}{UPC} & \multicolumn{11}{c}{Accuracy in each session ($\%$)}&\multirow{2}{*}{PD $\downarrow$} &\multirow{2}{*}{$Acc_{avg}$}\\
\cline{3-13}
& &0 & 1 & 2 & 3 & 4 & 5 & 6 & 7 & 8 & 9 & 10 \\ 
\hline
\thead{\normalsize \multirow{3}{*}[-2.0ex]{CIFAR100}} &\thead{30} &\thead{78.12} & \thead{72.23} & \thead{69.36} & \thead{64.91} & \thead{62.18} & \thead{59.67} & \thead{58.12} &\thead{56.02} & \thead{53.84} & \thead{-} & \thead{-} & \thead{24.28} &\thead{63.83} \\

&\thead{50} &\thead{78.52} & \thead{73.09} & \thead{70.37} & \thead{66.15} & \thead{63.94} & \thead{61.69} & \thead{59.91} &\thead{58.00} & \thead{55.95} & \thead{-} & \thead{-} & \thead{22.57} &\thead{65.29} \\

&\thead{70} &\thead{76.95} & \thead{71.75} & \thead{69.03} & \thead{64.51} & \thead{62.33} & \thead{59.91} & \thead{58.47} &\thead{56.44} & \thead{54.55} & \thead{-} & \thead{-} & \thead{22.40} &\thead{63.70}\\
\hline
\thead{\normalsize \multirow{3}{*}[-2.0ex]{{\it mini}ImageNet}}  &\thead{50} &\thead{75.72} & \thead{71.62} & \thead{68.11} & \thead{65.87} & \thead{63.59} & \thead{61.25} & \thead{59.24} &\thead{58.60} & \thead{57.84} & \thead{-} & \thead{-} & \thead{17.88} &\thead{64.65}\\

&\thead{70} &\thead{75.03} & \thead{71.11} & \thead{67.46} & \thead{65.77} & \thead{63.85} & \thead{61.35} & \thead{59.36} &\thead{59.21} & \thead{58.63} & \thead{-} & \thead{-} & \thead{16.40} &\thead{64.64}\\

&\thead{90} &\thead{74.97} & \thead{71.20} & \thead{67.94} & \thead{66.00} & \thead{64.48} & \thead{62.07} & \thead{59.99} &\thead{59.74} & \thead{58.99} & \thead{-} & \thead{-} & \thead{16.02} &\thead{65.04}\\

\hline
\thead{\normalsize \multirow{2}{*}[-1.6ex]{CUB200}} &\thead{5} &\thead{78.76} & \thead{74.70} & \thead{71.72} & \thead{66.85} & \thead{66.24} & \thead{63.03} & \thead{62.44} &\thead{60.22} & \thead{59.13} & \thead{58.59} & \thead{57.66} & \thead{21.10} &\thead{65.34}\\

&\thead{5-10} &\thead{78.30} & \thead{74.82} & \thead{71.90} & \thead{67.58} & \thead{66.83} & \thead{64.25} & \thead{62.92} &\thead{61.59} & \thead{60.65} & \thead{59.65} & \thead{58.69} & \thead{19.61} &\thead{66.11}\\

&\thead{10-15} &\thead{78.61} & \thead{74.95} & \thead{72.18} & \thead{67.82} & \thead{67.13} & \thead{64.17} & \thead{62.59} &\thead{61.16} & \thead{59.92} & \thead{59.14} & \thead{58.33} & \thead{20.28} &\thead{66.00}\\


\bottomrule
\label{tab:res7}
\end{tabular}}
\end{table*} 

    \textbf{Impact of pseudo-label quality} In the previous paper, we elucidated the theoretical aspects of how pseudo-label quality affects results and proposed approaches for generating pseudo-labels. To provide a more intuitive demonstration of the significance of pseudo-label quality, we devised a comparative experiment. Pseudo-labels were generated using various methods, and the disparities in the final results were observed. Detailed outcomes are presented in Table \ref{tab:res8}.
    
\begin{table*}[htbp]
\small 
\renewcommand\arraystretch{1.3}
\caption{IMPACT OF PSEUDO-LABEL QUALITY. WE SELECTED THE R18$\textsuperscript{†}$ AND RLCC FOR COMPARISON.}
\label{tab11}
\centering
\resizebox{\linewidth}{!}{\begin{tabular}{c c c c c c c c c c c c c c c c c c}
\toprule
\multicolumn{1}{c}{\multirow{2}{*}{Datasets}} &\multirow{2}{*}{Methods}  &\multirow{2}{*}{UPC} &\multirow{2}{*}{ARI} & \multicolumn{9}{c}{Accuracy in each session ($\%$)}&\multirow{2}{*}{PD $\downarrow$}  &\multirow{2}{*}{$Acc_{avg}$}\\
\cline{5-13}
& & & &0 & 1 & 2 & 3 & 4 & 5 & 6 & 7 & 8 \\ 
\hline
\thead{\normalsize \multirow{2}{*}[-1.4ex]{CIFAR100}}  &\thead{R18$\textsuperscript{†}$} &\thead{50} &\thead{0.12} &\thead{76.67} & \thead{70.80} & \thead{67.50} & \thead{63.79} & \thead{61.23} & \thead{58.59} & \thead{56.80} &\thead{54.74} & \thead{52.46} & \thead{24.21} &\thead{62.51} \\

&\thead{RLCC} &\thead{50} &\thead{0.30} &\thead{78.12} & \thead{72.23} & \thead{69.36} & \thead{64.91} & \thead{62.18} & \thead{59.67} & \thead{58.12} &\thead{56.02} & \thead{53.84} & \thead{24.28} &\thead{65.29} \\

\hline
\thead{\normalsize \multirow{2}{*}[-1.4ex]{{\it mini}ImageNet}}  &\thead{R18$\textsuperscript{†}$} &\thead{50} &\thead{0.21} &\thead{75.38} & \thead{71.19} & \thead{67.59} & \thead{65.47} & \thead{63.51} & \thead{61.04} & \thead{58.14} &\thead{57.08} & \thead{56.03} & \thead{19.35} &\thead{63.94}\\

&\thead{RLCC} &\thead{50} &\thead{0.38} &\thead{75.72} & \thead{71.62} & \thead{68.11} & \thead{65.87} & \thead{63.59} & \thead{61.25} & \thead{59.24} &\thead{58.60} & \thead{57.84} & \thead{17.88} &\thead{64.65} \\



\bottomrule
\label{tab:res8}
\end{tabular}}
\end{table*} 

\begin{figure}[htbp]
\centering
\includegraphics[height=6.2cm,width=8.5cm]{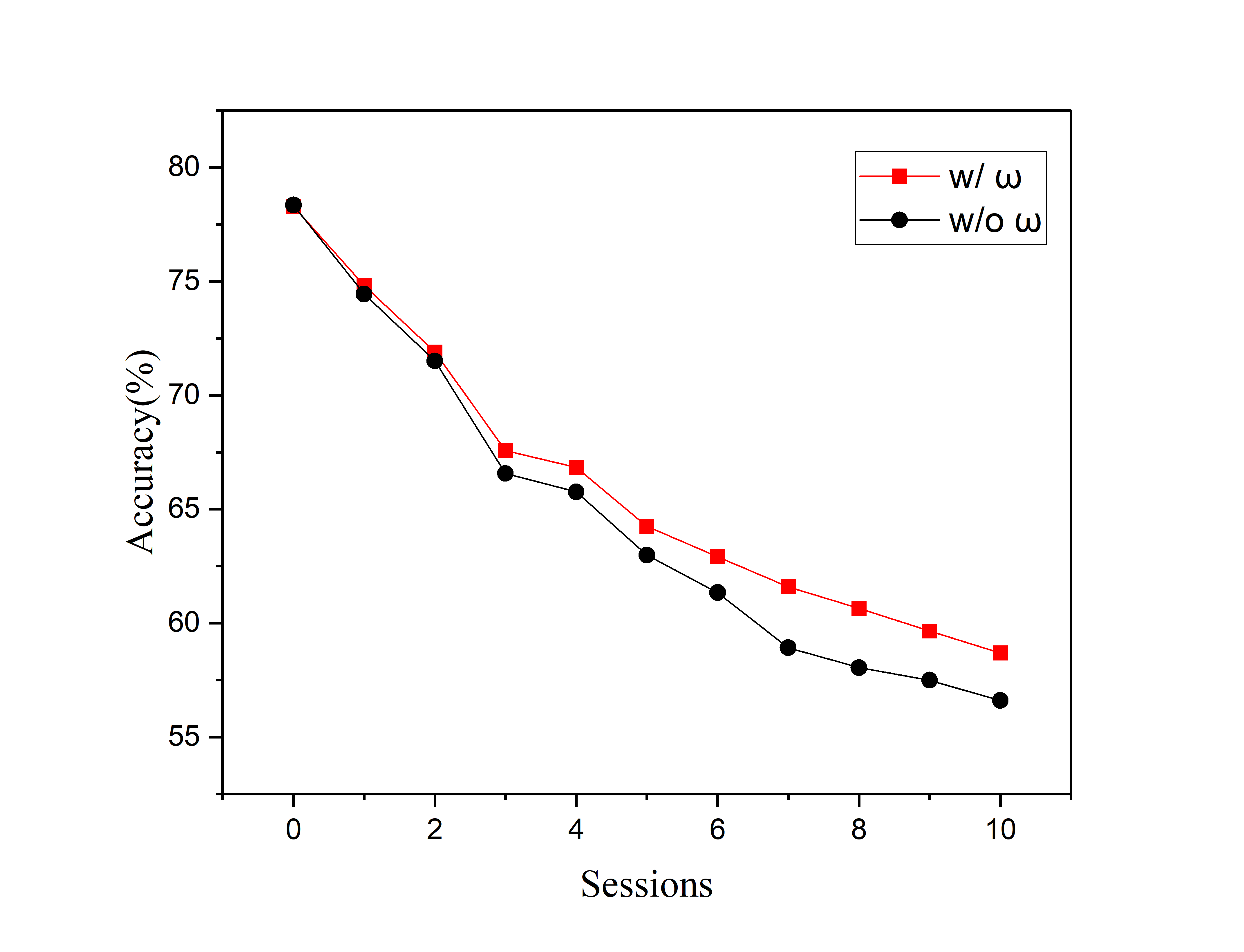}
\caption{Impact of the reconciliation coefficient $\omega$. ``w/" and ``w/o" denote with and without $\omega$, respectively.}
\label{fig:5}
\end{figure}

    \textbf{Impact of the reconciliation coefficient $\omega$} We introduce a reconciliation coefficient $\omega$ in the loss function to balance the weights of the base and new types of patterns in the model. To verify its effectiveness, we design relevant experiments on CUB200. The experimental results are shown in Figure \ref{fig:5}.
    
\subsection{Further Analysis}
In this study, we adopt a similar semi-supervised approach to enhance performance. Analysis of the experimental results indicates that the primary factors influencing the ultimate performance are the quantity and quality of the pseudo-labeled data. Significant improvements were achieved on the CIFAR100 and miniImageNet datasets, while the enhancement on CUB200 is less pronounced. CUB200, being a fine-grained dataset of birds, poses a more challenging learning task than the other two datasets. Its characteristic of more categories with fewer samples presents a substantial challenge to the RLCC module, thereby affecting the subsequent incremental learning phase. Notably, we demonstrate superior performance with a greater amount of unlabeled data. The RLCC module introduces additional parameters and training volume, but in return, we reduce the training volume in the incremental phase. Plans involve incorporating the RLCC module into the initial pre-training, eliminating the need for additional models, and enabling an end-to-end incremental learning paradigm.


\section{Conclusion}
In this work, we present the Learning with Prior Knowledge (LwPK) algorithm, demonstrating the value of incorporating prior knowledge into pre-trained models for Few-Shot Class-Incremental Learning (FSCIL). Data from incremental classes are labeled using clustering algorithm and co-trained with base class data during pre-training. This equips the model with essential prior knowledge for the incremental learning stage, while minimizing parameter adjustments to preserve previously acquired knowledge. Experimental results confirm the efficacy of LwPK, demonstrating its ability to advance in the incremental learning stage even with limited data. Additionally, the theoretical foundation of LwPK, based on empirical risk minimization and class distance analysis, corroborates the experimental observations. This study underscores the significance of prior knowledge in FSCIL and holds promise for guiding future solutions to similar challenges.
\bibliographystyle{IEEEtran}
\bibliography{ref}











\newpage

 





\end{document}